\pdfoutput=1

\documentclass[11pt]{article}

\usepackage{acl}

\usepackage{times}
\usepackage{latexsym}

\usepackage[T1]{fontenc}

\usepackage[utf8]{inputenc}

\usepackage{microtype}

\usepackage{inconsolata}
\usepackage{graphicx}
\usepackage{amsmath}
\usepackage{amsfonts}
\usepackage{booktabs}
\usepackage{float}
\usepackage{nicefrac}
\usepackage{color}
\usepackage{multirow}
\usepackage{array}

\newcommand{\p}{\mathrm{p}}
\newcommand{\PP}{\mathrm{P}}

\title{Handling Ambiguity in Emotion: \\
From Out-of-Domain Detection to Distribution Estimation
}

\author{Wen Wu$^1$, Bo Li$^2$,  Chao Zhang$^3$, Chung-Cheng Chiu$^2$, Qiujia Li$^2$, Junwen Bai$^2$, \\
{\bf Tara N. Sainath$^2$, Philip C. Woodland$^1$}\\
$^1$ University of Cambridge, UK, $^2$ Google, LLC, USA, $^3$ Tsinghua University, China \\
\small \texttt{$^1$\{ww368, pcw\}@eng.cam.ac.uk, $^2$\{boboli, chungchengc\}@google.com, $^3$cz277@tsinghua.edu.cn}
}

\begin{document}
\maketitle
\begin{abstract}

The subjective perception of emotion leads to  inconsistent labels from human annotators.
Typically, utterances lacking majority-agreed labels are excluded when training an emotion classifier, which cause problems when encountering ambiguous emotional expressions during testing.
This paper investigates three methods to handle ambiguous emotion.
First, we show that incorporating utterances without majority-agreed labels as an additional class in the classifier reduces the classification performance of the other emotion classes. 
Then, we propose detecting utterances with ambiguous emotions as out-of-domain samples by quantifying the uncertainty in emotion classification using evidential deep learning. This approach retains the classification accuracy while effectively detects ambiguous emotion expressions.
Furthermore, to obtain fine-grained distinctions among ambiguous emotions, we propose representing emotion as a distribution instead of a single class label. The task is thus re-framed from classification to distribution estimation where every individual annotation is taken into account, not just the majority opinion.
The evidential uncertainty measure is extended to quantify the uncertainty in emotion distribution estimation. 
Experimental results on the IEMOCAP and CREMA-D datasets demonstrate the superior capability of the proposed method in terms of majority class prediction, emotion distribution estimation, and uncertainty estimation\footnote{Code will be available upon acceptance.}. 
\end{abstract}

\section{Introduction}

The inherent subjectivity of human emotion perception introduces complexity in annotating  emotion datasets. Multiple annotators are often involved in labelling each utterance and the majority-agreed (MA) class is usually used as the ground truth~\cite{busso2008iemocap, cao2014crema}. Utterances that have no majority-agreed (NMA) labels (\emph{i.e.}, with tied votes) are typically excluded during emotion classifier training~\cite {Kim_2013,Poria2017,wu2021emotion}, which may cause issues when the system encounters such utterances in practical applications.

This paper investigates three approaches to handling ambiguous emotion data. First, a naive method is tested which aggregates NMA utterances into an additional class when training an emotion classifier. This approach proves problematic as NMA utterances contain a blend of emotions, thereby confusing the classifier and undermining the classification performance.

Then we explore if an emotion classifier can appropriately respond with ``I don't know'' for ambiguous emotion data that does not fit into any predefined emotion class. This is realised by quantifying the uncertainty in emotion classification using evidential deep learning (EDL)~\citep{sensoy2018evidential}. When a classifier trained on MA data encounters an NMA utterance during the test, the model should identify it as an out-of-domain (OOD) sample by providing a high uncertainty score, indicating its uncertainty regarding the specific emotion class to which the NMA utterance belongs.

\begin{figure}[tb]
    \centering
    \includegraphics[width=\linewidth]{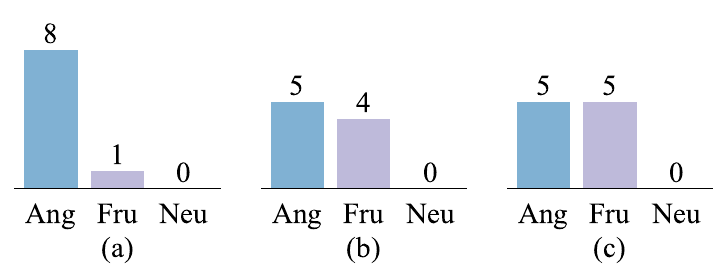}
    \vspace{-4ex}
    \caption{The bar chart shows the number of labels assigned by annotators to the emotion class ``angry'' (Ang), ``frustrated'' (Fru), and ``neutral'' (Neu) in an example. In utterance (a), eight annotators interpret the emotion as angry while one interprets it as frustrated.}
    \label{fig: eg}
\end{figure}

Moreover, to obtain fine-grained distinctions between ambiguous emotional data, we re-frame the task from classification to distribution estimation.
Consider the example shown in Figure~\ref{fig: eg} with the annotations assigned to three utterances.  Since the majority emotion classes are ``angry'' for both utterances (a) and (b), they will be assigned the same ground-truth label ``angry'' in the aforementioned classification system, which implies that they convey the same emotion content and is evidently unsuitable. On the contrary, utterance (c), though being an NMA utterance, is more likely to share similar emotional content with utterance (b).
Therefore, in order to obtain more comprehensive representations of emotion content, we further propose representing emotion as a distribution rather than a single class label and re-framing emotion recognition as a distribution estimation problem rather than a classification problem. A novel algorithm is proposed which extends EDL to  estimate the underlying emotion distribution given observed human annotations and quantify the uncertainty in emotion distribution estimation. 
The proposed approach considers all human annotations rather than relying solely on the majority vote class. 
Multiple evaluation metrics are adopted to evaluate the performance in terms of majority class prediction, uncertainty measure, and distribution estimation.
Rather than simply saying ``I don't know'', the proposed system demonstrates the ability to estimate the emotion distributions of the NMA utterances and also offer a reliable uncertainty measure for the distribution estimation.

Our contributions are summarised as follows. \textbf{(i)} 
To the best of our knowledge, this paper is the first work that treats ambiguous emotion as OOD and detects it by uncertainty estimation; \textbf{(ii)} This is the first work that applies EDL to quantify uncertainty in emotion classification; \textbf{(iii)} Imposing a single ground truth through majority voting leads to under-representation of minority views. We instead estimate the distribution over emotion classes which provides a more comprehensive representation of emotion content as well as a more inclusive representation of human opinions; \textbf{(iv)} A novel algorithm is proposed that extends EDL to quantify uncertainty in emotion distribution estimation.

\section{Related work}
\label{sec: literature}
Human annotators often interpret the emotion of the same utterance differently due to their personal experiences and cultural backgrounds~\cite{busso2008iemocap,cowen2017self,cummins2019ambiguous}. 
Instead of using the MA annotation as the ground truth label, some research suggests treating emotion classification as a multi-label task~\citep{mower2010framework,zadeh2018multimodal,chochlakis2023leveraging} where all emotion classes assigned by any annotator are considered as correct classes and the ground truth label is presented as a multi-hot vector. The model is trained to predict the presence of each emotion class for each utterance. %
An issue with this approach is that it ignores the differences in strengths of different emotion classes. 

An alternative approach uses ``soft labels'' as the proxy of ground truth, which is defined as the relative frequency of occurrence of each emotion class~\citep{Fayek_2016,han2017hard,kim2018human}. The Kullback–Leibler (KL) divergence or distance metrics between the soft labels and model predictions are used to train the model. However, soft labels, being maximum likelihood estimates (MLE) of the underlying distribution based on observed samples, might not provide an accurate approximation to the unknown distribution when the number of observations (annotations) is limited. Also, although adopting soft labels, those methods still focus on obtaining a ``correct'' label (\textit{i.e.,} pursuing improved classification accuracy).

So far, the calibration of emotion models has not been extensively studied. In this study, we introduce a novel approach which provides not only better emotion content estimation but also a reliable measure of the model's prediction confidence.

\section{Detecting NMA as OOD by quantifying emotion classification uncertainty}
\label{sec: method-ood}

As explained in the introduction and confirmed experimentally in Section~\ref{sec: MLE+}, 
training an emotion classifier with NMA utterances grouped into an additional class
degrades the classification performance. 
This section studies an alternative method. 
The emotion classifier is trained on MA utterances and NMA utterances are treated as OOD samples. By quantifying uncertainty in emotion classification, the model is expected to output a high uncertainty score when encountering ambiguous emotions, indicating that the utterance doesn't belong to any predefined MA class.

\subsection{Limitation of modelling class probabilities with the softmax activation function}
\label{sec: softmax}
A neural network model classifier transforms the continuous logits at the output layer into class probabilities by a softmax function. The model prediction can thus be interpreted as a categorical distribution with the discrete class probabilities associated with the model outputs. The model is then optimised by maximising the categorical likelihood of the correct class, known as the cross-entropy loss.

However, the softmax activation function is known to have a tendency to inflate the probability of the predicted class due to the exponentiation applied to transform the logits, resulting in unreliable uncertainty estimations~\cite{gal2016dropout,guo2017calibration}. 
Furthermore, cross-entropy is essentially MLE, a frequentist technique lacking the capability to infer the variance of the predictive distribution. 

In the following section, we estimate the model uncertainty using evidential deep learning (EDL)~\citep{sensoy2018evidential} which places a second-order probability over the categorical distribution.

\subsection{Quantify uncertainty in emotion classification by evidential deep learning}
\label{sec: method-edl}
Consider an emotion class label as a one-hot vector $\boldsymbol{y}$ where $y_k$ is one if the emotion belongs to class $k$ else zero. $\boldsymbol{y}$ is sampled from a categorical distribution $\boldsymbol{\eta}$ where each component ${\eta}_k$ corresponds to the probability of sampling a label from class $k$:
\begin{equation}
    \boldsymbol{y} \sim \mathrm{P}(\boldsymbol{y}|\boldsymbol{\eta}) = \operatorname{Cat}(\boldsymbol{\eta}) = {\eta}_k^{y_k}.
    \label{eqn: cat}
\end{equation}
To model the probability of the predictive distribution, we assume the categorical distribution is sampled from a Dirichlet distribution:
\begin{equation}
    \boldsymbol{\eta} \sim \mathrm{p}(\boldsymbol{\eta} | \boldsymbol{\alpha}) = \operatorname{Dir}(\boldsymbol{\eta} | \boldsymbol{\alpha}) = \frac{1}{B(\boldsymbol{\alpha})}\prod_{k=1}^K \eta_k^{\alpha_k-1}
\end{equation}
where $ B(\cdot)$ is the Beta function, ${\alpha}_k$ is the hyperparameter of the Dirichlet distribution. ${\alpha}_0= \sum_{k=1}^K \alpha_k$ is the Dirichlet strength. 
The output of a standard neural network classifier is a probability assignment over the possible classes and the Dirichlet distribution represents the probability of each such probability assignment, hence modelling second-order probabilities and uncertainty. %

Subjective logic~\citep{jsang2018subjective} establishes a connection between the Dirichlet distribution and the belief representation in Dempster–Shafer belief theory~\citep{dempster1968generalization}, also known as evidence theory.  
Consider $K$ classes each associated with a belief mass $b_k$ and an overall uncertainty mass $u$, which satisfies $u+\sum_{k=1}^K b_k =1$.
The belief mass assignment corresponds to the Dirichlet hyperparameter $\alpha_k$: $b_k = {(\alpha_k-1)}/{{\alpha_0}}$, where
$e_k=\alpha_k-1$ is usually termed evidence. The overall uncertainty can then be computed as: 
\begin{equation}
    u=\frac{K}{{\alpha_0}}.
    \label{eqn: uncertainty}
\end{equation}

A neural network $\boldsymbol{f_\Lambda}$ is trained to predict $\operatorname{Dir}(\boldsymbol{\eta}^{(i)} | \boldsymbol{\alpha}^{(i)})$ for a given sample $\boldsymbol{x}^{(i)}$ where $\boldsymbol{\Lambda}$ is the model parameters. The network is similar to standard neural networks for classification except that the softmax output layer is replaced with a ReLU activation layer to assure non-negative outputs, which is taken as the evidence vector for the predicted Dirichlet distribution: $\boldsymbol{f_\Lambda} (\boldsymbol{x}^{(i)}) = \boldsymbol{e}^{(i)}$. The concentration parameter of the Dirichlet distribution can be calculated as $\boldsymbol{\alpha}^{(i)}= \boldsymbol{f_\Lambda} (\boldsymbol{x}^{(i)}) +1$.
Given $\operatorname{Dir}(\boldsymbol{\eta}^{(i)} | \boldsymbol{\alpha}^{(i)})$, the estimated probability of class $k$ can be calculated by:
\begin{equation}
    \mathbb{E}[{\eta}^{(i)}_k]=\frac{{\alpha}^{(i)}_k}{{\alpha_0}^{(i)}}.
\end{equation}

\subsubsection{Training}
For brevity, superscript $i$ is omitted in this section. Given one-hot label $\boldsymbol{y}$ and predicted Dirichlet $\operatorname{Dir}(\boldsymbol{\eta} | \boldsymbol{\alpha})$, the network can be trained by maximising the marginal likelihood of sampling $\boldsymbol{y}$ given the Dirichlet prior. Since the Dirichlet distribution is the conjugate prior of the categorical distribution, the marginal likelihood is tractable:
\begin{equation}
    \begin{aligned}
    \PP(\boldsymbol{y}|\boldsymbol{\alpha}) &= \int \PP(\boldsymbol{y}|\boldsymbol{\eta}) \p(\boldsymbol{\eta}|\boldsymbol{\alpha}) d\boldsymbol{\eta} \\
    &=\int \prod_k {\eta}_k^{y_k} \frac{1}{B(\boldsymbol{\alpha})}\prod_k {\eta}_k^{{\alpha}_k-1} \\
    &=\frac{B(\boldsymbol{\alpha}+\boldsymbol{y})}{B(\boldsymbol{\alpha})}     = \frac{\prod_{k=1}^K{\alpha}_k^{{y}_k}}{{\alpha_0}^{\sum_{k=1}^K{y}_k}}.
    \label{eqn: ood-marginal}
\end{aligned}
\end{equation}
It is equivalent to training the model by minimising the negative log marginal likelihood:
\begin{align}
    \mathcal{L}^{\text{NLL}} =  \sum_{k=1}^K{y}_k (\log({\alpha}_0) - \log({\alpha_k})).
    \label{eqn: NLL loss}
\end{align}
Following~\citep{sensoy2018evidential}, a regularisation term is added to penalise the misleading evidence:
\begin{equation}
    \mathcal{L}^\text{R} =  \mathcal{KL}(\operatorname{Dir}(\boldsymbol{\eta}|\boldsymbol{\Tilde{\alpha}})||\operatorname{Dir}(\boldsymbol{\eta}|\boldsymbol{1})),
    \label{eqn: reg-original}
\end{equation}
where $\operatorname{Dir}(\boldsymbol{\eta}| \boldsymbol{1})$ denotes a Dirichlet distribution with zero total evidence and $\boldsymbol{\Tilde{\alpha}} = \boldsymbol{y} + (1-\boldsymbol{y})\odot \boldsymbol{\alpha}$ is the Dirichlet parameters after removal of the non-misleading evidence from predicted $\boldsymbol{\alpha}$. This penalty explicitly enforces the total evidence to shrink to zero for a sample if it cannot be correctly classified.
The overall loss is $ \mathcal{L} = \mathcal{L}^{\text{NLL}}+\lambda \mathcal{L}^\text{R}$
where $\lambda$ is the regularisation coefficient.

\section{Emotion distribution estimation}
\label{sec: method-distribution}
As illustrated in Figure~\ref{fig: eg}, the majority vote class is not sufficient for fine-grained emotion representations. In this section, we describe emotion by a distribution instead of a single class label.

Consider an input utterance $\boldsymbol{x}^{(i)}$ associated with $M_i$ labels from human annotators $\{\boldsymbol{y}_m^{(i)}\}_{m=1}^{M_i}$ where $\boldsymbol{y}_m=[y_{m1},\ldots, y_{mK}]$ is a one-hot vector. Instead of representing the emotion content by the majority vote class, we propose estimating the underlying emotion distribution $\boldsymbol{\eta}$ based on the observations $\{\boldsymbol{y}_m^{(i)}\}_{m=1}^{M_i}$. The emotion classification problem is thus re-framed as a distribution estimation problem.
In contrast to the ``soft label'' method in Section~\ref{sec: literature} which approximates the emotion distribution of each $\boldsymbol{x}^{(i)}$ solely based on $\mathcal{D}^{(i)}=\{\boldsymbol{y}_m^{(i)}\}_{m=1}^{M_i}$ by MLE and trains the model to learn this proxy in a supervised manner, the proposed approach meta-learns a distribution estimator $\boldsymbol{f_\Lambda}$ across all data points $\mathcal{D}_\text{meta}=\{\mathcal{D}^{(i)}\}_{i=1}^N$ where $N$ is the number of utterances in training. This uses the knowledge about the emotion expression and annotation variability across different utterances.

For brevity, superscript $i$ is omitted thereafter. Assume $\{\boldsymbol{y}_m\}_{m=1}^M$ are samples drawn from a multinomial distribution. 
Let $\hat{\boldsymbol{y}}=\sum_{m=1}^M \boldsymbol{y}_{m}$ represent the counts of each emotion class:
\begin{align}
    \{\boldsymbol{y}_m\}_{m=1}^M \sim \mathrm{P}(\boldsymbol{y}|\boldsymbol{\eta}) =\operatorname{Mult}(\boldsymbol{\eta}, M)\\
    \operatorname{Mult}(\boldsymbol{\eta}, M) = \frac{\Gamma(M+1)}{\prod_{k=1}^K\Gamma(\hat{y}_k+1)}{\eta}_k^{\hat{y}_k}.
    \label{eqn: multinomial}
\end{align}
The categorical distribution in Eqn.~\eqref{eqn: cat} is the special case when $M=1$.

The network is trained by maximising the marginal likelihood of sampling $\{\boldsymbol{y}_m\}_{m=1}^M$ given the predicted Dirichlet prior $\operatorname{Dir}(\boldsymbol{\eta} | \boldsymbol{\alpha})$:
    \begin{align}
    \nonumber &\PP(\{\boldsymbol{y}_m\}_{m=1}^M|\boldsymbol{\alpha}) = \int \PP(\{\boldsymbol{y}_m\}_{m=1}^M|\boldsymbol{\eta}) \p(\boldsymbol{\eta}|\boldsymbol{\alpha}) d\boldsymbol{\eta} \\
    &= \frac{\Gamma(M+1)}{\prod_{k=1}^K\Gamma(\hat{y}_k+1)} \frac{\prod_{k=1}^K{\alpha}_k^{\hat{y}_k}}{{\alpha_0}^{\sum_{k=1}^K\hat{y}_k}}.
    \label{eqn: ood-marginal}
\end{align}
The multinomial coefficient is independent of $\boldsymbol{\alpha}$, we thus verify that $\mathcal{L}^{\text{NLL}}$ in Eqn.~\eqref{eqn: NLL loss} can be generalised to the distribution estimation framework by replacing one-hot majority label $\boldsymbol{y}$ with $\hat{\boldsymbol{y}}$:
\begin{align}
    \mathcal{L}^{\text{NLL}^*} =  \sum_{k=1}^K\hat{y}_k (\log({\alpha}_0) - \log({\alpha_k})).
    \label{eqn: NLL loss}
\end{align}

The regulariser in Eqn.~\eqref{eqn: reg-original} is then replaced with:
\begin{equation}
    \mathcal{L}^\text{R1}=\mathcal{KL}(\operatorname{Dir}(\boldsymbol{\eta}|\boldsymbol{\hat{\alpha}})||\operatorname{Dir}(\boldsymbol{\eta}|\boldsymbol{1}))
    \label{eqn: R1}
\end{equation}
where $\boldsymbol{\hat{\alpha}} = \boldsymbol{\bar{y}} + (1-\boldsymbol{\bar{y}})\odot \boldsymbol{\alpha}$ and $\bar{\boldsymbol{y}}=\frac{1}{M}\sum_{m=1}^M \boldsymbol{y}_{m}$ is the soft label. An alternative regulariser is proposed in order to explicitly regularise the predicted multinomial distribution:
\begin{equation}
     \mathcal{L}^\text{R2}=\mathcal{KL}(\bar{\boldsymbol{y}}||\mathbb{E}[{\boldsymbol{\eta}}]).
     \label{eqn: KL}
\end{equation}
Hence, we have extend the EDL method described in Section~\ref{sec: method-edl} for classification to quantify the uncertainty in distribution estimation, with the original method~\citep{sensoy2018evidential} being a special case when $M=1$ and $\hat{\boldsymbol{y}}$ becomes the one-hot majority label $\boldsymbol{y}$.
In addition, it's worth noting that the proposed approach does not require a fixed number of annotators for every utterance and can easily generalise to a large number of annotators (\emph{i.e.}, for crowd-sourced datasets).

\section{Evaluation metrics}
\label{sec: metric}
The proposed method is evaluated in terms of majority prediction, uncertainty estimation, OOD detection, and distribution estimation.

\textbf{Majority prediction.}  Majority prediction for MA utterances is evaluated by classification accuracy (ACC) and unweighted average recall (UAR) which is the sum of class-wise accuracy divided by the number of classes. 

\textbf{Uncertainty estimation.} Model calibration is evaluated by expected calibration error (ECE)~\citep{naeini2015obtaining} and maximum calibration error (MCE)~\citep{naeini2015obtaining}. ECE measures model calibration by computing the difference in expectation between confidence and accuracy. Predictions are partitioned into Q bins equally spaced in the [0,1] range and ECE can be computed as follows:
\begin{equation}
    \text{ECE} = \sum_{q=1}^Q \frac{|B_q|}{n}\left| \text{Acc}(B_q) - \text{Conf}(B_q)\right|.
\end{equation}
MCE is a variation of ECE which measures the largest calibration gap:
\begin{equation}
    \text{MCE} = \max_{q\in\{1,\ldots,Q\}} \left| \text{Acc}(B_q) - \text{Conf}(B_q)\right|.
\end{equation}

\textbf{OOD detection.} The area under the receiver operating characteristic (AUROC) and the area under the precision-recall curve (AUPRC) are used to evaluate the performance of OOD detection. 
The estimated uncertainty is used as a decision threshold for both AUROC and AUPRC. The baseline is 50\% for AUROC and is the fraction of positives for AUPRC. NMA utterances are set as the positive class to detect.

\textbf{Distribution estimation.}  Emotion distribution estimation performance is measured by the negative log-likelihood (NLL) of sampling human annotations from the predicted multinomial distribution.

\section{Experimental setup}
\label{sec: exp setup}

\subsection{Baselines}
The proposed methods were compared to the following baselines:
\vspace{-1ex}
\begin{itemize}
    \item MLE: a deterministic classification network with softmax activation trained by the cross-entropy loss between the majority vote label and model predictions;
    \vspace{-1ex}
    \item MCDP: a Monte-Carlo dropout~\citep{gal2016dropout} model with a dropout rate of 0.5 which is forwarded 100 times to obtain 100 samples during testing;
    \vspace{-1ex}
    \item Ensemble: an ensemble~\citep{lakshminarayanan2017simple} of 10 MLE models with the same structure trained by bagging;
    \vspace{-1ex}
    \item MLE+: a MLE model with NMA as an extra class.
    \vspace{-1ex}
\end{itemize}
An additional baseline for distribution estimation:
\begin{itemize}
\vspace{-1ex}
    \item MLE*: the ``soft label'' approach mentioned in Section~\ref{sec: literature} which is trained by minimising KL divergence between the soft label $\bar{\boldsymbol{y}}$ and predictions. It is an extension of MLE from one-hot majority vote labels to soft labels.
    \vspace{-0.5ex}
\end{itemize}

The system described in Section~\ref{sec: method-edl} is denoted as \textbf{``EDL''}. \textbf{``EDL*(R1)''} and \textbf{``EDL*(R2)''}  refers to the systems proposed in in Section~\ref{sec: method-distribution} using regularisation terms defined in Eqn.~\eqref{eqn: R1} and Eqn.~\eqref{eqn: KL} respectively. Uncertainty estimation of EDL models are computed by Eqn.~\eqref{eqn: uncertainty} while max probability is used as confidence measure for other methods.

\begin{table*}[t]
\centering
\resizebox{\linewidth}{!}{%
\begin{tabular}{c|cccc|cc|cc}
\toprule
 & \multicolumn{4}{c|}{\textbf{Classify MA}}& \multicolumn{2}{c|}{\textbf{Detect NMA (all)}}& \multicolumn{2}{c}{\textbf{Detect NMA (test)}}\\
 & \textbf{ACC $\uparrow$}& \textbf{UAR $\uparrow$}& \textbf{ECE $\downarrow$} & \textbf{MCE $\downarrow$} & \textbf{AUROC $\uparrow$} & \textbf{AUPRC $\uparrow$} & \textbf{AUROC $\uparrow$} & \textbf{AUPRC $\uparrow$} \\
 \midrule
MLE+ & 0.447 & 0.438 & 0.303 & 0.383 & / & / & 0.461 & 0.139 \\
MLE & 0.582 & 0.577 & 0.206 & 0.239 & 0.550 & 0.471 & 0.549 & 0.177 \\
MCDP & 0.584 & 0.572 & \underline{0.128} & \underline{0.184} & 0.566 & \underline{0.491} & \underline{0.568} & \underline{0.203} \\
Ensemble & \underline{0.593} & \underline{0.595} & 0.439 & 0.594 & \underline{0.567} & \underline{0.491} & 0.563 & 0.192 \\
EDL & \textbf{0.611} & \textbf{0.596} & \textbf{0.103} & \textbf{0.145} & \textbf{0.610} & \textbf{0.530} & \textbf{0.620} & \textbf{0.227} \\
\bottomrule
\end{tabular}
}
\vspace{-1ex}
\caption{Results of quantifying uncertainty in emotion classification on the IEMOCAP dataset. The baseline for AUPRC is 0.433 for the entire NMA set and 0.160 for the NMA test subset. The best value in each column is indicated in bold, and the second-best value is underlined.}
\label{tab: ood-iemo}
\end{table*}

\begin{table*}[t]
\centering
\resizebox{\linewidth}{!}{%
\begin{tabular}{c|cccc|cc|cc}
\toprule
 & \multicolumn{4}{c|}{\textbf{Classify MA}}& \multicolumn{2}{c|}{\textbf{Detect NMA (all)}}& \multicolumn{2}{c}{\textbf{Detect NMA (test)}}\\
 & \textbf{ACC $\uparrow$}& \textbf{UAR $\uparrow$}& \textbf{ECE $\downarrow$} & \textbf{MCE $\downarrow$} & \textbf{AUROC $\uparrow$} & \textbf{AUPRC $\uparrow$} & \textbf{AUROC $\uparrow$} & \textbf{AUPRC $\uparrow$} \\
 \midrule
MLE+ & 0.568 & 0.540 & 0.216 & 0.476 & / & / & 0.552 & 0.156 \\
MLE & 0.714 & 0.672 & 0.150 & 0.156 & 0.578 & 0.467 & 0.571 & 0.179 \\
MCDP & \underline{0.717} & \underline{0.687} & \underline{0.102} & \underline{0.109} & \underline{0.619} & \underline{0.481} & \underline{0.614} & \underline{0.201} \\
Ensemble & \textbf{0.731} & 0.674 & 0.362 & 0.496 & 0.598 & \underline{0.481} & 0.605 & 0.198 \\
EDL & 0.711 & \textbf{0.714} & \textbf{0.057} & \textbf{0.080} & \textbf{0.645} & \textbf{0.506} & \textbf{0.657} & \textbf{0.234} \\
\bottomrule
\end{tabular}
}
\vspace{-1ex}
\caption{Results of quantifying uncertainty in emotion classification on the CREMA-D dataset. The baseline for AUPRC is 0.387 for the entire NMA set and 0.097 for the NMA test subset.}
\label{tab: ood-cremad}
\end{table*}

\subsection{Datasets}
Two publicly available datasets were used in the experiments. 
The IEMOCAP corpus~\citep{busso2008iemocap} is one of the most widely used emotion datasets. It consists of 10,039 English utterances from 5 dyadic conversational sessions. Each utterance was evaluated by at least three human annotators.
Only 16.1\% of utterances have an all-annotators-agreed emotion label.
The emotion distribution is represented using a five-dimensional categorical distribution, including happy (merged with excited), sad, neutral, angry, and others. The ``others'' category includes all emotions not covered in the previous four categories which is dominated by frustration (92\%). 14.2\% of the utterances don't have a majority agreed emotion class label.

The CREMA-D corpus~\citep{cao2014crema} contains 7,442 English utterances from 91 actors. Actors spoke from a selection of 12 sentences using one of six different emotions (anger, disgust, fear, happy, neutral and sad). The dataset was annotated by crowd-sourcing. Ratings based on audio alone were used in this work.
Utterances have 9.21 ratings on average. 5.1\% of utterances have an all-annotators-agreed emotion label and 8.7\%  don't have a majority agreed emotion class label. 

Both datasets were divided into an MA subset and an NMA subset. 
All methods were trained only on MA data except for MLE+ where 25\% of NMA utterances were reserved for testing and the rest were included in training. For IEMOCAP, Session 5 was reserved for testing, and Sessions 1-4 were split into training and validation with a ratio of 4:1.
For the CREMA-D dataset, the MA subset was split into train, validation, test in the ratio 70 : 15 : 15 following \citet{ristea21_interspeech}.

\begin{figure}[t]
    \centering
    \includegraphics[width=\linewidth]{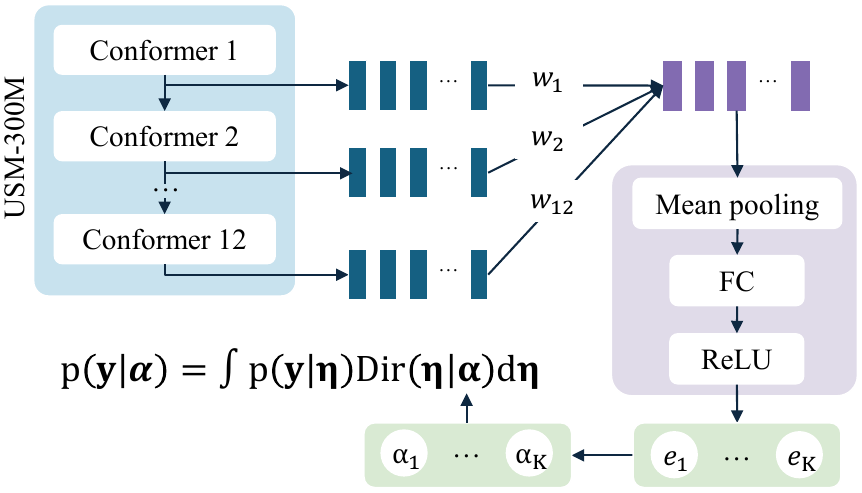}
    \caption{Illustration of the model structure.}
    \vspace{-1ex}
    \label{fig: struc}
\end{figure}
\subsection{Model structure}
The backbone structure used in this paper is illustrated in Figure~\ref{fig: struc} which follows an upstream-downstream paradigm~\citep{bommasani2021opportunities}. The upstream model uses the universal speech model (USM)~\citep{zhang2023google} with 300M parameters which contains a CNN-based feature extractor and 12 Conformer~\cite{gulati20_interspeech} encoder blocks of dimension 1024 with 8 attention heads. 
The structure of the downstream model follows  SUPERB~\citep{yang21c_interspeech}, a benchmark for evaluating pre-trained upstream models, which performs utterance-level mean-pooling followed by a fully-connected layer. 
The pre-trained upstream USM model is frozen. The downstream model computes the weighted sum of the hidden states extracted from each layer of the upstream model. The backbone structure has been shown to outperform state-of-the-art methods for emotion classification (see Table~\ref{tab: superb} in Appendix~\ref{apdx: superb}). The implementation details for model training can be found in Appendix~\ref{apdx: implementation}.

\section{Results}
\label{sec: results}
This section presents experimental results of the three approaches for handling ambiguous emotion: incorporating NMA as an extra class (Section~\ref{sec: MLE+}), detecting NMA as OOD (Section~\ref{sec: exp ood}), and representing emotion as distributions (Section~\ref{sec: exp distr}). The average of three runs are reported for all results.

\subsection{Including NMA as an additional category degrades the performance} 
\label{sec: MLE+}

The first approach, which trains an emotion classifier with NMA as an extra class, is denoted ``MLE+'' in Table~\ref{tab: ood-iemo} and ~\ref{tab: ood-cremad}. Some of the NMA utterances are included in MLE+ training while the remainder are used for testing. Therefore, OOD detection is evaluated only on NMA (test) data for MLE+. 
The results reveal that the addition of the NMA  class has a detrimental impact on the classification performance of the original MA emotion classes. Comparing to MLE, MLE+ observes a $\sim$23\% relative decrease in both ACC and UAR on IEMOCAP and a $\sim$20\% relative decrease in ACC and UAR on CREMA-D. The confusion matrices of the MLE+ model can be found in Appendix~\ref{apdx: confusion}, which shows that NMA itself is challenging to predict and it also confuses the model when predicting classes such as neutral, sad, frustrated, and disgust.

\subsection{Detecting NMA as OOD}
\label{sec: exp ood}
The proposed EDL-based method is compared to baselines in Tables~\ref{tab: ood-iemo} and ~\ref{tab: ood-cremad} on the IEMOCAP and CREMA-D datasets respectively.
First, as shown by the values of ACC and UAR, the proposed method demonstrates comparable classification performance to the baselines, suggesting that the extension for uncertainty estimation does not undermine the model's capabilities. Although the Ensemble achieves the highest accuracy on CREMA-D, it involves training 10 individual systems. The proposed method achieves overall the best classification performance with only a tenth of the computational cost of Ensemble during both training and testing. 
In addition, the proposed method offers superior model calibration, as shown by the lowest values of ECE and MCE. 
It also outperforms the baselines in effectively identifying NMA as OOD samples, as shown by the highest AUROC and AUPRC values.

\begin{figure}[htb]
    \centering
    \includegraphics[width=0.95\linewidth]{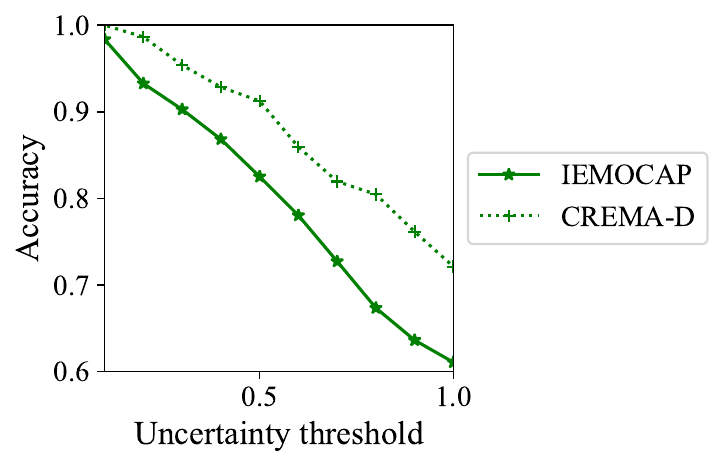}
    \vspace{-1ex}
    \caption{The change of accuracy with respect to
the uncertainty threshold for EDL-based methods on IEMOCAP and CREMA-D.}
    \label{fig: rej acc both}
\end{figure}

Figure~\ref{fig: rej acc both} shows the change of accuracy when samples with uncertainty larger than a threshold are excluded. 
The model tends to provide less accurate predictions when it is less confident about its prediction, shown by the decrease of classification accuracy when the uncertainty threshold increases, which demonstrates the effectiveness of uncertainty prediction.

\subsection{Estimating emotion distribution}
\label{sec: exp distr}
\begin{table}[b]
\centering
\begin{tabular}{c|cccc}
\toprule
\textbf{IEMOCAP} &\textbf{ACC}& \textbf{UAR }& \textbf{ECE } & \textbf{MCE } \\
 \midrule
MLE* & 0.564 & 0.562 & 0.151 & 0.279 \\
EDL*(R1) & 0.623 & 0.612 & 0.081 & 0.208  \\
EDL*(R2) & \textbf{0.624} & \textbf{0.616} & \textbf{0.025} & \textbf{0.201} \\
\midrule
\midrule
\textbf{CREMA-D}&\textbf{ACC}& \textbf{UAR }& \textbf{ECE } & \textbf{MCE } \\
 \midrule
MLE* & 0.693 & 0.621 & 0.109 & 0.115 \\
EDL*(R1) & \textbf{0.740} & 0.694 & \textbf{0.029} & \textbf{0.095} \\
EDL*(R2) & 0.718 & \textbf{0.722} & 0.084 & 0.107 \\
\bottomrule
\end{tabular}
\vspace{-1ex}
\caption{Classification and calibration performance of distribution-based methods on MA data. The best value in each column is indicated in bold.}
\label{tab: distri-MA}
\end{table}

The proposed EDL* methods were first evaluated in terms of majority class prediction.
The results of distribution-based methods on classification of MA data are shown in Table~\ref{tab: distri-MA}. Compared to the classification-based methods in Table~\ref{tab: ood-iemo} and Table~\ref{tab: ood-cremad}, it can be seen that EDL* does not reduce the performance of emotion classification (in terms of ACC and UAR) and model calibration (in terms of ECE and NCE) on MA data. This indicates that the information of the majority class is retained when representing emotion as a distribution.
Note that when representing emotion as a distribution, it is no longer appropriate to consider NMA utterances as OOD samples, as illustrated by case (b) and (c) in Figure~\ref{fig: eg}. 
Although still trained only on MA data, the proposed distribution-based system shows good generalisation ability in predicting the emotion distribution of NMA data, which we will see shortly.  When encountering NMA data during testing, instead of simply returning ``I don't know'', the proposed system can provide reliable estimation of its emotional content, which is a key benefit.

The proposed EDL* methods were then evaluated regarding distribution estimation.
Table~\ref{tab: nll} compares EDL* to the baselines in terms of the negative log likelihood of sampling target labels from the predicted emotion distribution. As can be seen from the table, EDL* produce improved distribution estimation, achieving the smallest NLL values on both MA and NMA data. Among the two EDL* methods employing different regularisation terms, EDL* with R2 (defined in Eqn.~\eqref{eqn: KL}), which directly applies regularisation to the predicted distribution, exhibits better distribution estimation without sacrificing model calibration.

\begin{table}[tb]
\centering
\begin{tabular}{c|cc}
\toprule
\textbf{IEMOCAP} & \textbf{NLL$^{\text{MA}} \downarrow$}& \textbf{NLL$^{\text{NMA}} \downarrow$} \\
 \midrule
MLE & 1.310 & 1.924 \\
MCDP & 0.972 & 1.266 \\
Ensemble & 2.572 & 2.055 \\
EDL & 0.958 & 1.019 \\
\midrule
MLE* & 0.941 & 1.137 \\
EDL*(R1) & 0.861 & \textbf{0.951} \\
EDL*(R2) & \textbf{0.833} & 0.953 \\

\midrule
\midrule
\textbf{CREMA-D}& \textbf{NLL$^{\text{MA}} \downarrow$}& \textbf{NLL$^{\text{NMA}} \downarrow$} \\
 \midrule
MLE & 1.532 & 2.054 \\
MCDP & 0.965 & 1.292 \\
Ensemble & 2.285 & 2.089 \\
EDL & 0.757 & 1.021 \\
\midrule
MLE* & 0.648 & 0.774 \\
EDL*(R1) & 0.614 & 0.722 \\
EDL*(R2) & \textbf{0.606} & \textbf{0.698} \\
\bottomrule
\end{tabular}
\caption{Distribution estimation results. NMA stands for NMA(all).}
\vspace{-0.5ex}
\label{tab: nll}
\end{table}

\begin{figure}[tb]
    \centering
    \includegraphics[width=0.9\linewidth]{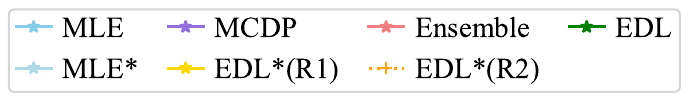}
    \vskip 0.5ex
    \begin{minipage}[b]{0.47\linewidth}
\small\centerline{\includegraphics[width=\linewidth]{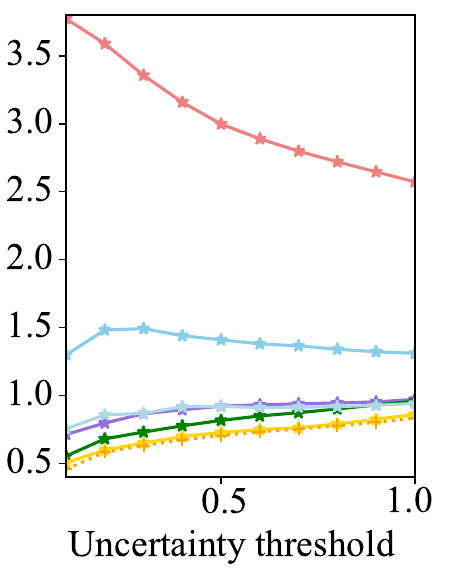}}
\vskip 0.5ex
  \centerline{(a) MA data}
    \end{minipage}
    \begin{minipage}[b]{0.47\linewidth}
\small\centerline{\includegraphics[width=\linewidth]{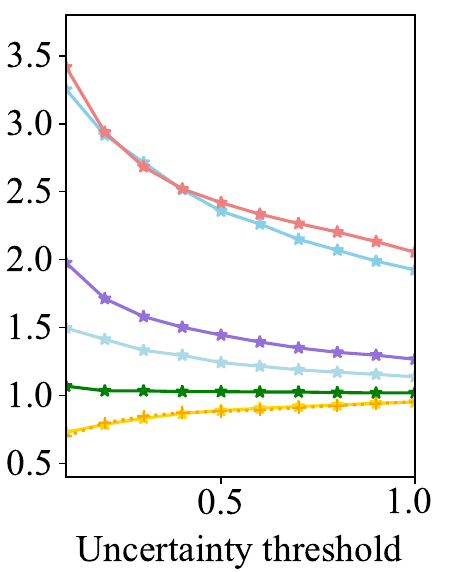}}
\vskip 0.5ex
  \centerline{(b) NMA data}
    \end{minipage}
        \caption{Reject option for NLL on IEMOCAP. Trends on CREMA-D are similar, shown in Appendix~\ref{apdx: rej-cremad}.}
    \label{fig: NLL-rej-iemo}
\end{figure}

A reject option was then evaluated for NLL (instead of accuracy) to examine model calibration. For a well-calibrated model, an increase in the NLL value, which is associated with poorer distribution estimation, is expected when the model becomes less confident. 
Figure~\ref{fig: NLL-rej-iemo} visualises the change of NLL for MA data and NMA data when uncertainty increases. 
For MA data, \emph{i.e.}  the type of data that has been seen by the models during training, most methods can successfully reject uncertain samples except for MLE and Ensemble, as shown by an increase in NLL values when the uncertainty threshold increases. However, for NMA data which the model hasn't seen in training, only the EDL* methods exhibit the ability to demonstrate an increasing trend in NLL values.

The proficiency of the proposed EDL* methods for estimating the emotion distribution and providing reliable confidence predictions, demonstrates the method's capacity to estimate both aleatoric uncertainty~\citep{matthies2007quantifying,der2009aleatory}, arising from data complexity (\emph{i.e.}, the ambiguity of emotion expression), and epistemic uncertainty, corresponding to the amount of uncommitted belief in subjective logic.

\subsection{Case study}
\label{sec: case study}
\begin{figure}[tb]
    \centering
    \includegraphics[width=\linewidth]{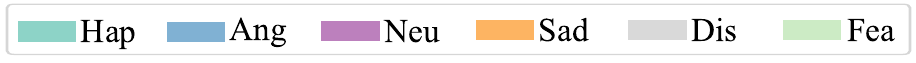}
    \vskip 0.5ex
    \includegraphics[width=0.49\linewidth]{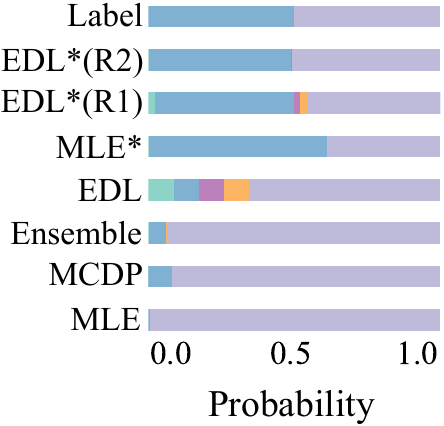}
    \includegraphics[width=0.49\linewidth]{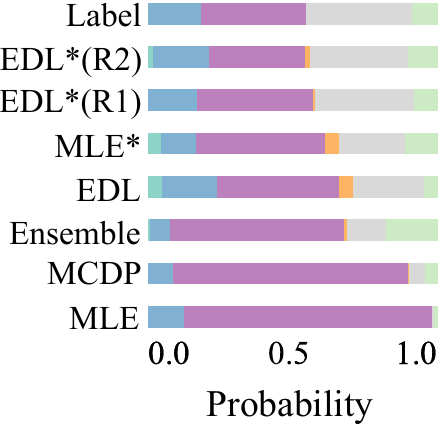}
    \begin{minipage}[b]{0.49\linewidth}
  \small \centerline{(a) ``Ses04M\_impro02\_F024''}
    \end{minipage}
    \begin{minipage}[b]{0.49\linewidth}
  \small \centerline{(b) ``1084\_TSI\_ANG\_XX''}
    \end{minipage}
    \caption{Visualisation of emotion distribution for case study. Utterance (a)  from IEMOCAP. Utterance (b) from CREMA-D.}
    \label{fig: case study}
\end{figure}

Emotion distributions estimated by different methods are visualised against the label distributions for two representative examples in Figure~\ref{fig: case study}. In general, distribution-based methods show superior performance for distribution estimation than classification-based methods. In the case of  utterance (a) which receives two ``angry'' labels and two ``frustrated'' labels, the proposed EDL* methods stands out by effectively capturing the tie between the emotions, whereas the predictions of classification-based methods tend to be predominantly skewed towards ``frustrated''. As for utterance (b), where both ``disgust'' and ``neutral'' receive four votes, along with two votes for ``angry'' and one for ``fear'', the emotion distributions predicted by the EDL* methods also show a similar pattern. These examples show that the proposed method can not only provide a more comprehensive emotion representation but also better reflect the variability of human opinions. Additional examples can be found in  Appendix~\ref{apdx: case-iemo} and Appendix~\ref{apdx: case-cremad}.

\section{Conclusions}
This paper re-examines the emotion classification problem, starting with an exploration of ways to handle data with ambiguous emotions. 
It is first shown that incorporating ambiguous emotions as an extra class reduces the classification performance of the original emotion classes. 
Then, evidence theory is adopted to quantify uncertainty in emotion classification which allows the classifier to output ``I don't know'' when it encounters utterances with ambiguous emotion. 
The model is trained to predict the hyperparameters of a Dirichlet distribution, which models the second-order probability of the probability assignment over emotion classes. 
Furthermore, to capture finer-grained emotion differences, we transform the emotion classification problem into an emotion distribution estimation problem. All annotations are taken into account rather than only the majority opinion. A novel approach is proposed which extends standard EDL to quantify uncertainty in emotion distribution estimation. Experimental results show that given an utterance with ambiguous emotion the proposed approach is able to provide a comprehensive representation of its emotion content as a distribution with a reliable uncertainty measure.

\section*{Ethics Statement}
In this work, all human annotations used for training were taken from existing publicly available corpora. No new human annotations were collected. 

In subjective tasks like emotion recognition, there is usually no single ``correct'' answer. The conventional approach of imposing a single ground truth through majority voting may overlook valuable nuances within each annotator's evaluation and the disagreements between them, potentially resulting in the under-representation of minority views. This study, instead of exclusively relying on the majority vote, integrates emotion annotations from all annotators for each utterance during model training. It is hoped that this work could contribute to a more inclusive representation of human opinions.

\section*{Limitations}
The proposed approach requires the raw labels from different human annotators for each sentence to be provided by the datasets. 
Although validated only for emotion recognition, the proposed method could also be applied to other tasks with disagreements in subjective annotations, which will be investigated in future work.

Different people perceive emotion differently and hence the motivation of the paper is to handle such ambiguity. It may be that if annotators were to also provide confidence ratings during annotation, which is not the case in the emotion datasets we have used, then this information could be used to weight the observations when estimating the emotion distribution.

\bibliography{anthology,custom}
\newpage
\appendix

\section{Implementation details}
\label{apdx: implementation}
This section describes the implementation details. The model is implemented using Pax\footnote{https://github.com/google/paxml}.
The batch size is set to 256, The coefficient $\lambda$ is set to 0.8 for IEMOCAP and 0.2 for CREMA-D. The Adafactor optimiser and Noam learning rate scheduler are used with 200 warm up steps and a peak learning rate of 8.84$\times 10^{-4}$. Since the CREMA-D dataset is extremely imbalanced (\emph{i.e.}, neutral accounts for over 50\%), a balanced sampler is applied during training.
The model is trained for 20k steps which takes $\sim$ 5 hours on 8 TPU v4s. 

\section{Discussion of statistical significance}
The results reported in Section~\ref{sec: results} are the average of three runs with different seeds. Regarding ECE, MCE, AUROC and AUPRC, the improvements in the results in bold are consistent across all three runs. For ACC and NLL, the differences are statistically significant with p < 0.05.

\section{Comparing the backbone structure to SOTA models}
\label{apdx: superb}
The USM-based backbone structure is evaluated following the setup of the emotion recognition task of the SUPERB benchmark~\citep{yang21c_interspeech}: four-way emotion classification (happy, sad, angry, neutral) on IEMOCAP dataset with leave-one-session-out five-fold cross validation. 
The USM-300M model is compared to multiple state-of-the-art models of similar size.
Results are shown in Table~\ref{tab: superb}. Except for the USM-300M model used in the paper, all other results are quoted from the cited papers. As shown in the table, the USM-based backbone structure outperforms other state-of-the-art methods\footnote{https://superbbenchmark.org/leaderboard} and yields the highest accuracy.

\begin{table}[htb]
    \centering
    \begin{tabular}{ccc}
    \toprule
         \textbf{Model}&  \textbf{\# Param} &\textbf{ ACC (\%)}\\
         \midrule
         \begin{tabular}[c]{@{}c@{}}Wav2vec 2.0 large\\ \citep{baevski2020wav2vec}\end{tabular} &   317M& 65.64\\
         
         \begin{tabular}[c]{@{}c@{}}Data2vec large\\ \citep{baevski2022data2vec}\end{tabular} &   314M& 66.31\\
         
         \begin{tabular}[c]{@{}c@{}}HuBERT large\\ \citep{hsu2021hubert}\end{tabular} &   317M & 67.62\\
         \begin{tabular}[c]{@{}c@{}}WavLM large\\ \citep{chen2022wavlm}\end{tabular} &  317M & 70.62\\
         \begin{tabular}[c]{@{}c@{}}USM-300M\\ \citep{zhang2023google}\end{tabular}
         &   290M& \textbf{71.06}\\
         \bottomrule
    \end{tabular}
    \caption{Four-way classification results IEMOCAP following the SUPERB-ER benchmark setup.}
    \label{tab: superb}
\end{table}

\section{Analysis of confusion matrices of MLE+}
\label{apdx: confusion}
\begin{figure}[H]
    \centering
    \begin{minipage}[b]{0.7\linewidth}
\centerline{\includegraphics[width=\linewidth]{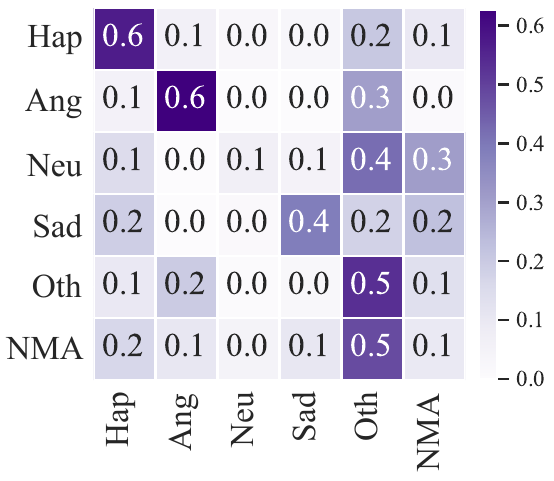}}
  \small\centerline{(a) IEMOCAP}
    \end{minipage}
    \vskip 2ex
    \begin{minipage}[b]{0.75\linewidth}
\centerline{\includegraphics[width=\linewidth]{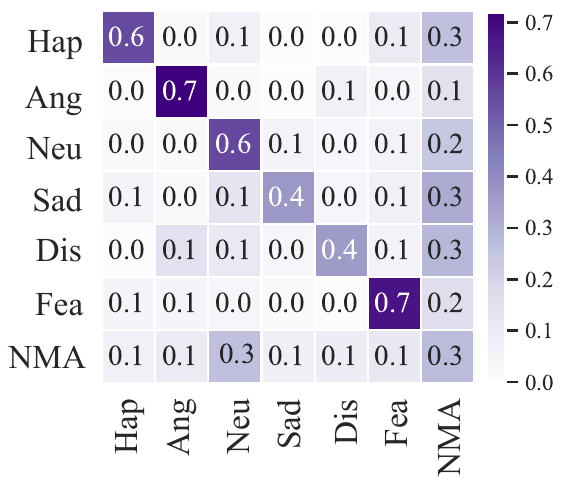}}
  \small\centerline{(b) CREMA-D}
    \end{minipage}
    \caption{Confusion matrix of the MLE+ system on IEMOCAP and CREMA-D.}
    \label{fig: confusion}
\end{figure}

As described in Section~\ref{sec: MLE+}, including NMA as an extra class reduces the classification performance.  This section analyses the confusion matrices of the MLE+ model, shown in Figure~\ref{fig: confusion}. It can be seen from the bottom right entry that NMA itself is challenging to predict, possibly because it  essentially contains a mix of different emotion content. The last column demonstrates that grouping these utterances into one class can confuse the model, particularly for the classes neutral, sad, frustrated, and disgust.

\begin{table*}[tb]
\centering
\resizebox{\linewidth}{!}{%
\begin{tabular}{c|cccc|cc|cc}
\toprule
  & \multicolumn{4}{c|}{\textbf{Classify MA}}& \multicolumn{2}{c|}{\textbf{Detect NMA (all)}}& \multicolumn{2}{c}{\textbf{Detect NMA (test)}}\\
 \textbf{IEMOCAP}& \textbf{ACC}& \textbf{UAR }& \textbf{ECE } & \textbf{MCE } & \textbf{AUROC } & \textbf{AUPRC } & \textbf{AUROC} & \textbf{AUPRC} \\
 \midrule
EDL (ReLU) & \textbf{0.611} & 0.596 & 0.103 & \textbf{0.145} & 0.610 & 0.530 & 0.620 & 0.227 \\
EDL (Softplus) & 0.608 & 0.574 & \textbf{0.035} & 0.173 & \textbf{0.617} & \textbf{0.534} & \textbf{0.639} & \textbf{0.251} \\
EDL (Exponential) & 0.588 & \textbf{0.601} & 0.167 & 0.230 & 0.593 & 0.502 & 0.619 & 0.225 \\
\midrule
\midrule

  & \multicolumn{4}{c|}{\textbf{Classify MA}}& \multicolumn{2}{c|}{\textbf{Detect NMA (all)}}& \multicolumn{2}{c}{\textbf{Detect NMA (test)}}\\
 \textbf{CREMA-D}& \textbf{ACC}& \textbf{UAR }& \textbf{ECE } & \textbf{MCE } & \textbf{AUROC } & \textbf{AUPRC } & \textbf{AUROC} & \textbf{AUPRC} \\
 \midrule
EDL (ReLU) & 0.701 & \textbf{0.714} & \textbf{0.057} & \textbf{0.080} & \textbf{0.645} & \textbf{0.506} & \textbf{0.657} & \textbf{0.234} \\
EDL (Softplus) & 0.692 & 0.696 & 0.113 & 0.309 & 0.640 & \textbf{0.506} & 0.633 & 0.230 \\
EDL (Exponential) & \textbf{0.723} & 0.602 & 0.277 & 0.277 & 0.623 & 0.495 & 0.626 & 0.197 \\
\bottomrule
\end{tabular}
}
\caption{Comparison of EDL methods with different activation functions on IEMOCAP and CREMA-D.}
\label{tab: activation-cremad}
\end{table*}
\section{Alternative activation functions}
\label{sec:appendix}

\begin{figure}[htb]
        \centering
        \includegraphics[width=0.48\linewidth]{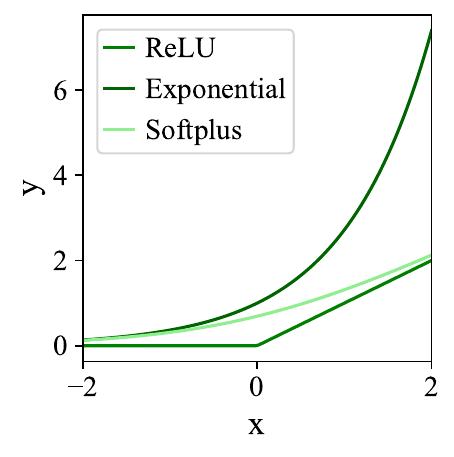}
        \vspace{-2ex}
        \caption{Illustration of the activation functions.}
        \label{fig: activation}
    \end{figure}
As described in Section~\ref{sec: method-edl}, ReLU is used as the output activation function in EDL to make sure the evidence is non-negative. This section compares the use of different activation functions including ReLU, softplus and exponential functions. The three activation functions are plotted in Figure~\ref{fig: activation}. 

\begin{figure}[H]
        \centering
        \includegraphics[width=0.7\linewidth]{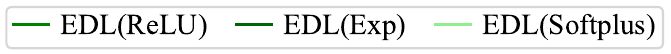}
        \begin{minipage}[b]{0.49\linewidth}
        \centerline{\includegraphics[width=\linewidth]{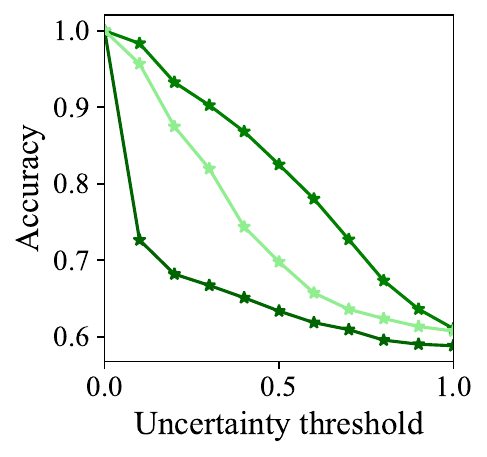}}
          \small\centerline{(a) IEMOCAP}
        \end{minipage}
        \begin{minipage}[b]{0.49\linewidth}
        \centerline{\includegraphics[width=\linewidth]{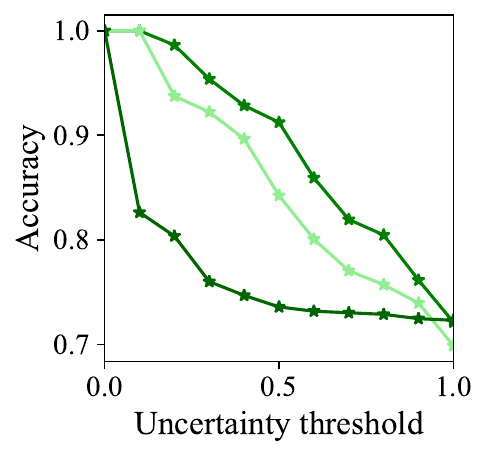}}
          \small\centerline{(b) CREMA-D}
        \end{minipage}
        \vspace{-3ex}
        \caption{Reject option for accuracy for EDL methods with different activation functions.}
        \label{fig: reject-both}
    \end{figure}

As shown in Table~\ref{tab: activation-cremad},  using exponential function tends to result in less effective model calibration, shown by the largest ECE and MCE values. It also produces worse performance for NMA detection, shown by the smallest AUROC and AUPRC. Figure~\ref{fig: reject-both} shows the reject option for accuracy of EDL with different activation functions. A drop in accuracy when the uncertainty threshold increases from 0 to 0.1 is observed for model using exponential activation. This indicates that exponential activation tends to lead to smaller uncertainty.

\begin{figure}[htb]
        \centering
        \includegraphics[width=0.7\linewidth]{fig/legend-activation-nobg.pdf}
        \begin{minipage}[b]{0.49\linewidth}
    \centerline{\includegraphics[width=\linewidth]{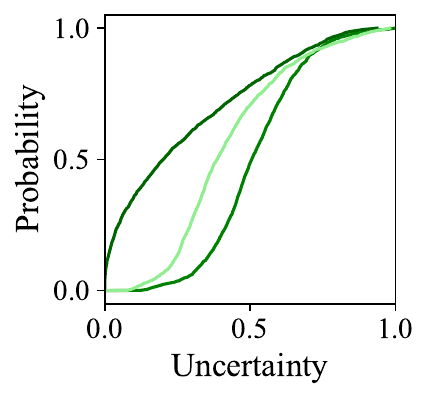}}
          \small\centerline{(a) ECDF of uncertainty}
        \end{minipage}
        \begin{minipage}[b]{0.49\linewidth}
    \centerline{\includegraphics[width=\linewidth]{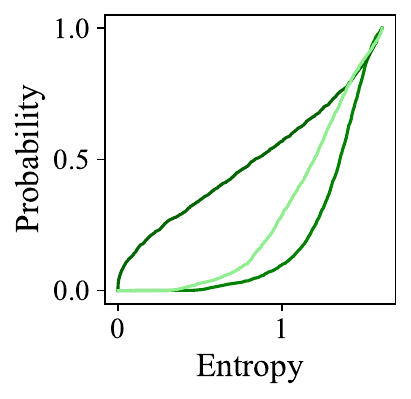}}
          \small\centerline{(b) ECDF of entropy}
        \end{minipage}
        \caption{Empirical CDF of uncertainty (left) and entropy (right) on IEMOCAP for EDL method with different activation functions.}
        \label{fig: activation-iemo}
    \end{figure}

\begin{figure}[H]
        \centering
        \includegraphics[width=0.8\linewidth]{fig/legend-activation-nobg.pdf}
        \begin{minipage}[b]{0.49\linewidth}
        \centerline{\includegraphics[width=\linewidth]{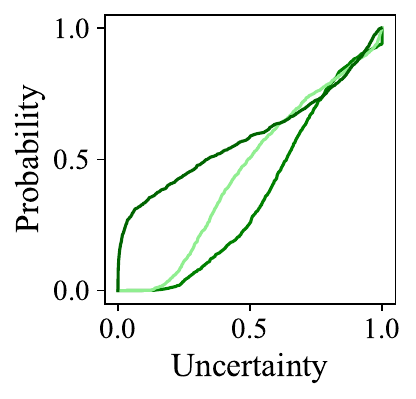}}
          \small\centerline{(a) ECDF of uncertainty}
        \end{minipage}
        \begin{minipage}[b]{0.49\linewidth}
        \centerline{\includegraphics[width=\linewidth]{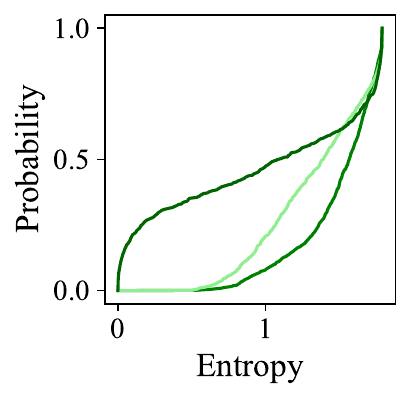}}
          \small\centerline{(b) ECDF of entropy}
        \end{minipage}
        \caption{Empirical CDF of uncertainty (left) and entropy (right) on CREMA-D for EDL method with different activation functions.}
        \label{fig: activation-cremad}
    \end{figure}

The empirical cumulative distribution function (ECDF) of uncertainty and entropy on IEMOCAP and CREMA-D are plotted in Figure~\ref{fig: activation-iemo} and Figure~\ref{fig: activation-cremad} respectively. It can be seen that exponential activation leads to smaller uncertainty and entropy, which echos the statement in Section~\ref{sec: softmax} that exponential activation tends to inflate the probability of the correct class.

\newpage
\section{Analysis of regularisation coefficient}
\begin{figure}[h]
        \centering
        \begin{minipage}[b]{0.45\linewidth}
    \centerline{\includegraphics[width=\linewidth]{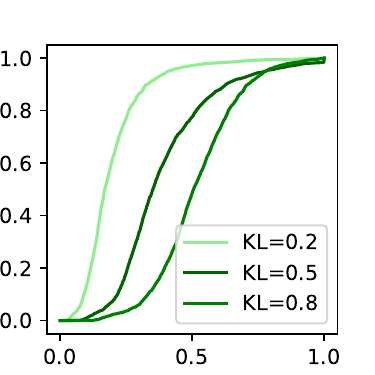}}
          \small\centerline{(a) ECDF of uncertainty}
        \end{minipage}
        \begin{minipage}[b]{0.45\linewidth}
    \centerline{\includegraphics[width=\linewidth]{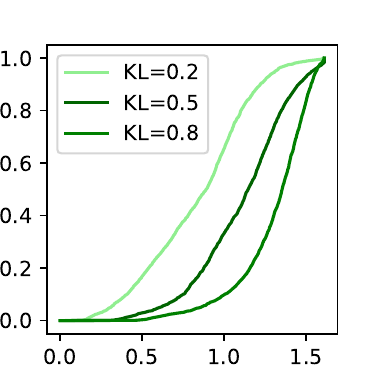}}
          \small\centerline{(b) ECDF of entropy}
        \end{minipage}
        \caption{Empirical CDF of uncertainty (left) and entropy (right) on IEMOCAP for EDL method with different regularisation coefficient $\lambda$.}
        \label{fig: lambda-iemocap}
        \vspace{-3ex}
    \end{figure}

\begin{figure}[h]
        \centering
        \begin{minipage}[b]{0.45\linewidth}
        \centerline{\includegraphics[width=\linewidth]{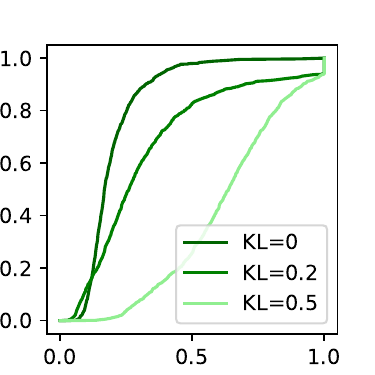}}
          \small\centerline{(a) ECDF of uncertainty}
        \end{minipage}
        \begin{minipage}[b]{0.45\linewidth}
        \centerline{\includegraphics[width=\linewidth]{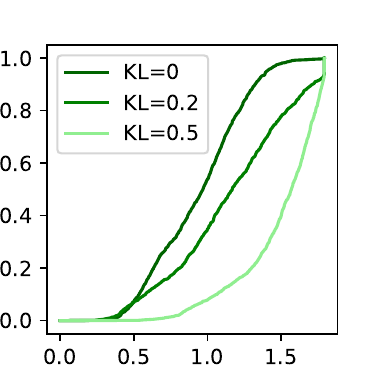}}
          \small\centerline{(b) ECDF of entropy}
        \end{minipage}
        \caption{Empirical CDF of uncertainty (left) and entropy (right) on CREMA-D for EDL method with different regularisation coefficient $\lambda$.}
        \vspace{-1ex}
        \label{fig: lambda-cremad}
    \end{figure}

This section analyses the effect of regularisation coefficient $\lambda$ in Eqn.~\eqref{eqn: reg-original} for EDL. The empirical CDF of uncertainty and entropy when different regularisation coefficient was used is plotted in in Figure~\ref{fig: lambda-iemocap} and Figure~\ref{fig: lambda-cremad} for IEMOCAP and CREMA-D respectively. We observed that larger lambda values lead to a larger entropy and uncertainty. This aligns with the definition of the regularisation term in Eqn.~\eqref{eqn: reg-original} which tends to enforce a flat prior with small evidence.

\newpage
\section{Reject option for NLL on CREMA-D}
\begin{figure}[H]
    \centering
    \includegraphics[width=0.7\linewidth]{fig/legend-nobg.pdf}
    \includegraphics[width=\linewidth]{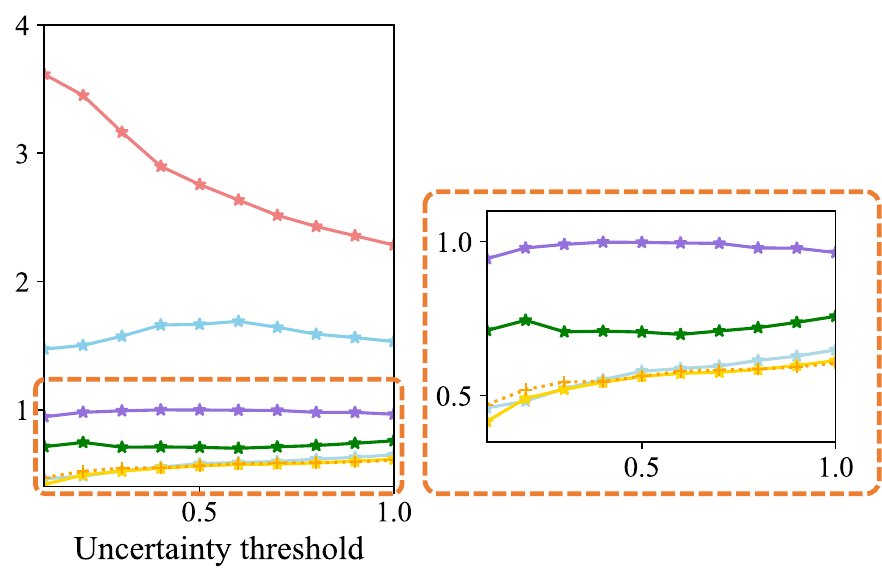}
    \caption{Reject option for NLL on MA data of CREMA-D.}
    \label{fig: reject nll cremad ma}
\end{figure}

This section shows the reject option for NLL on CREMA-D dataset. The change of NLL for MA data and NMA data when the uncertainty threshold increases are shown in Figure~\ref{fig: reject nll cremad ma} and Figure~\ref{fig: reject nll cremad ma} respectively. 
For a well-calibrated model, an increase in the NLL value is expected when the model becomes less confident. 
Similar to the findings in Figure~\ref{fig: NLL-rej-iemo}, most methods are effective for rejecting uncertain samples in the MA data, as shown by an increase in NLL values when the uncertainty threshold increases. However, only the EDL* methods are successful for NMA data.
\label{apdx: rej-cremad}
\begin{figure}[H]
    \centering
    \includegraphics[width=0.7\linewidth]{fig/legend-nobg.pdf}
    \includegraphics[width=\linewidth]{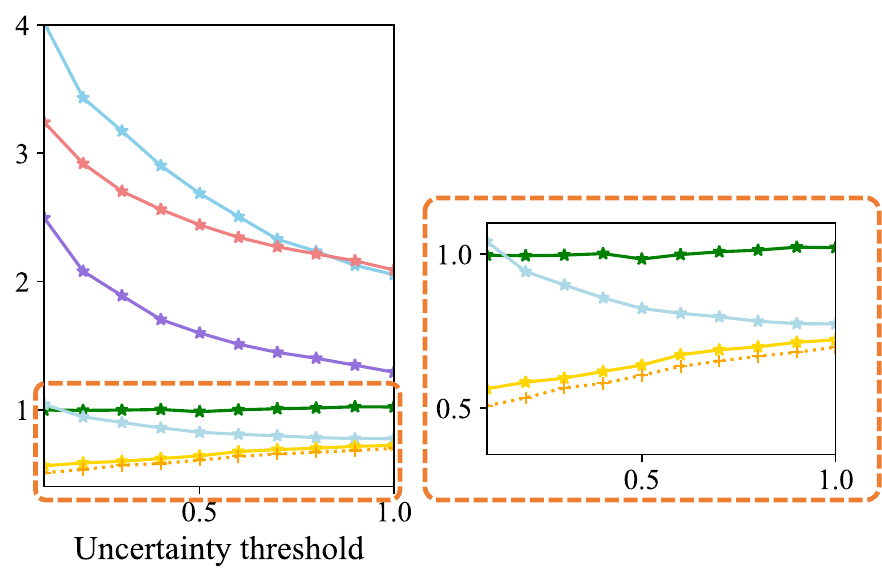}
    \caption{Reject option for NLL on NMA data of CREMA-D.}
    \label{fig: reject nll cremad nma}
\end{figure}

\newpage
\section{Further visualised examples: IEMOCAP}
\label{apdx: case-iemo}
This section shows more examples on IEMOCAP. Aligning with the findings in Section~\ref{sec: case study}, EDL* methods show better estimation of emotion distribution.
\begin{figure}[H]
    \centering
    \begin{minipage}[b]{0.8\linewidth}
        \centerline{\includegraphics[width=\linewidth]{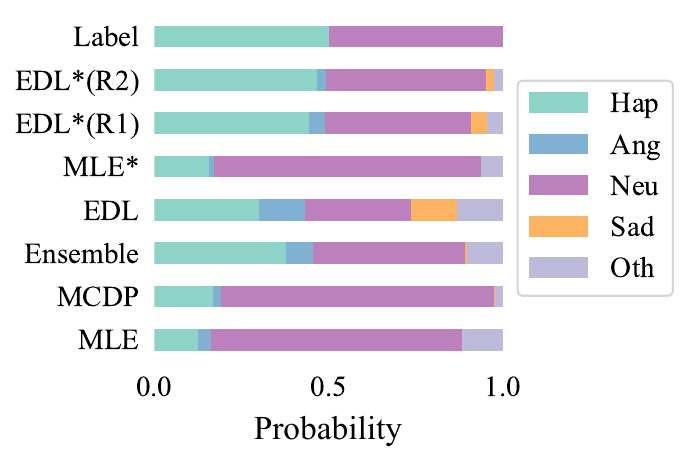}}
      \small\centerline{(a) Ses01M\_impro07\_M025}
    \end{minipage}
    \begin{minipage}[b]{0.8\linewidth}
        \centerline{\includegraphics[width=\linewidth]{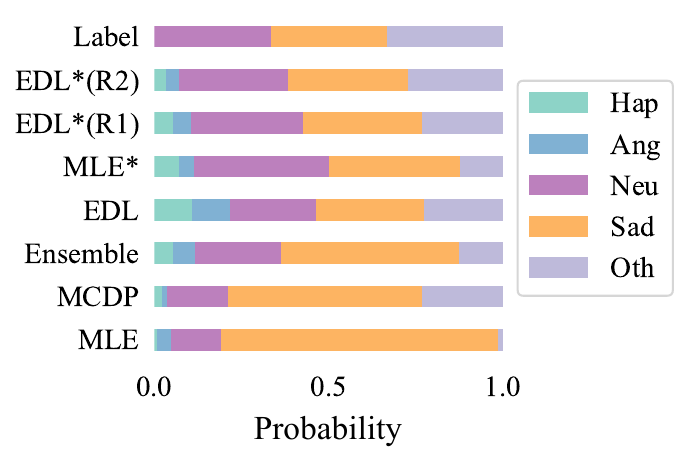}}
      \small\centerline{(b) Ses03F\_script01\_1\_F016}
    \end{minipage}
    \begin{minipage}[b]{0.8\linewidth}
        \centerline{\includegraphics[width=\linewidth]{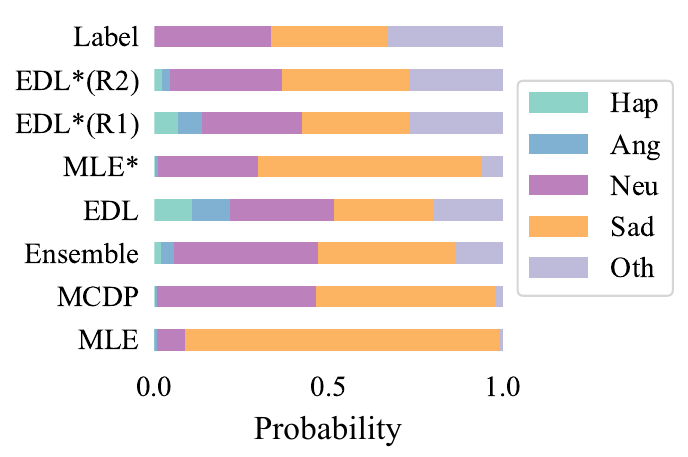}}
      \small\centerline{(c) Ses04F\_script01\_1\_M033}
    \end{minipage}
    \begin{minipage}[b]{0.8\linewidth}
        \centerline{\includegraphics[width=\linewidth]{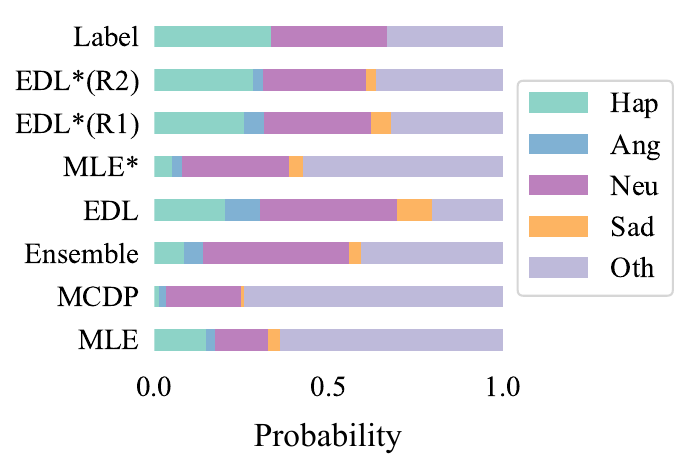}}
      \small\centerline{(d) Ses04M\_script01\_3\_M013}
    \end{minipage}
    \caption{Case study on IEMOCAP.}
\end{figure}

\newpage
\section{Further visualised examples: CREMA-D}
This section shows more examples of CREMA-D. As can be seen, EDL* methods can better approximate the distribution of emotional content of an utterance.
\label{apdx: case-cremad}
\begin{figure}[H]
    \centering
    \begin{minipage}[b]{0.8\linewidth}
        \centerline{\includegraphics[width=\linewidth]{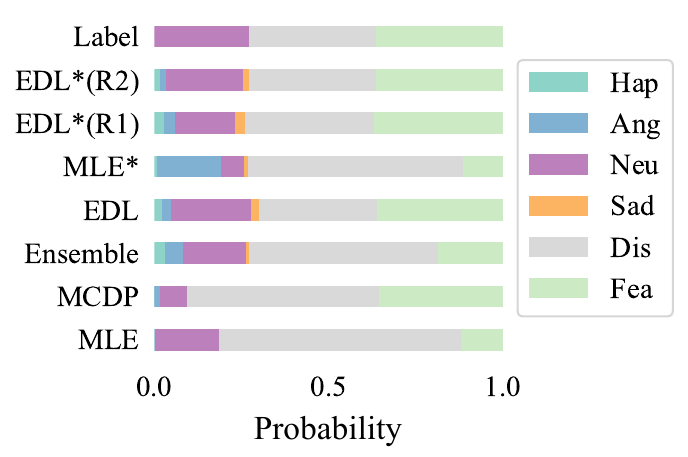}}
      \small\centerline{(a) 1033\_IWW\_DIS\_XX}
    \end{minipage}
    \begin{minipage}[b]{0.8\linewidth}
        \centerline{\includegraphics[width=\linewidth]{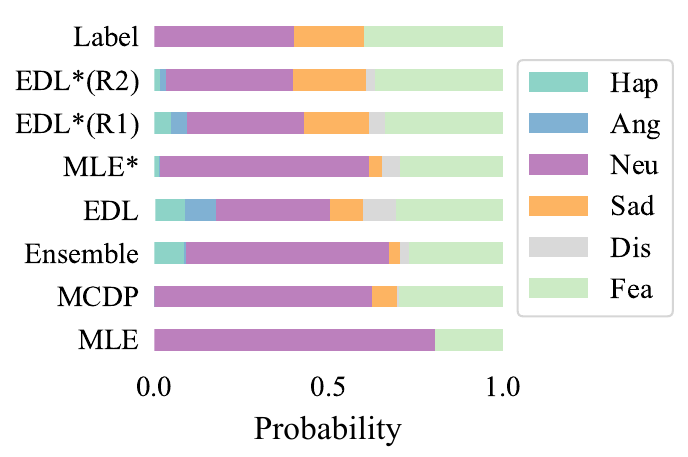}}
      \small\centerline{(b) 1052\_ITH\_FEA\_XX}
    \end{minipage}
    \begin{minipage}[b]{0.8\linewidth}
        \centerline{\includegraphics[width=\linewidth]{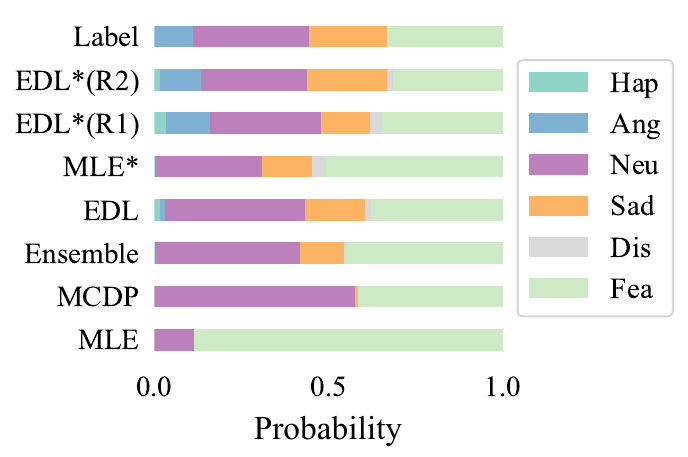}}
      \small\centerline{(c) 1068\_ITH\_SAD\_XX}
    \end{minipage}
    \begin{minipage}[b]{0.8\linewidth}
        \centerline{\includegraphics[width=\linewidth]{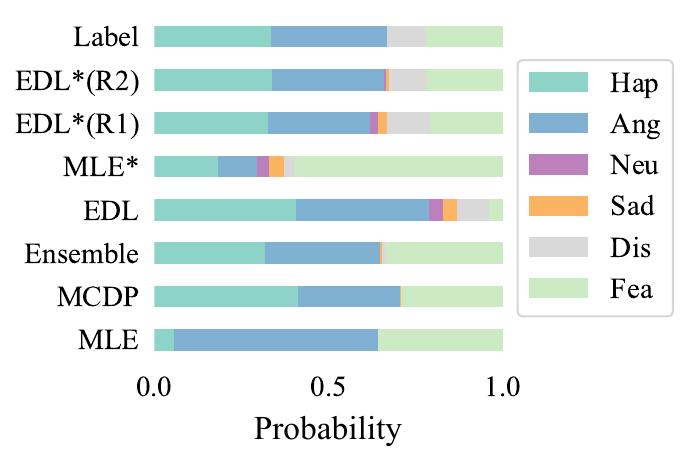}}
      \small\centerline{(d) 1009\_IWL\_FEA\_XX}
    \end{minipage}
    \caption{Case study on CREMA-D.}
    \label{fig: cremad-case}
\end{figure}

\newpage
\section{Analysis of Problematic Instances for OOD Detection}
This section includes examples and analysis of particular utterances where OOD detection is problematic and compares these examples across the techniques discussed.
It is shown that distribution-based methods improve over the classification-based systems in handling complex ambiguous emotions.

\begin{figure}[H]
    \centering
    \vspace{0.5ex}
    \includegraphics[width=0.6\linewidth]{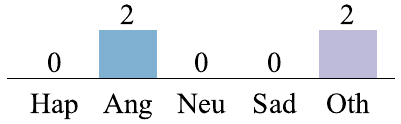}
    \vspace{-1ex}
    \caption{Human annotations for NMA utterance ``Ses04M\_impro02\_F024''.}
    \label{fig: iemo-22}
\end{figure}

\begin{figure}[H]
    \centering
    \includegraphics[width=0.6\linewidth]{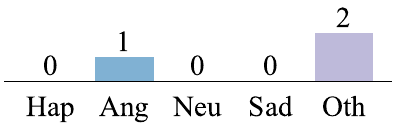}
    \vspace{-1ex}
    \caption{Human annotations for MA utterance ``Ses05M\_impro01\_M014''.}
    \label{fig: iemo-12}
\end{figure}

Consider the following false negative case where the OOD detection model fails to detect an NMA sample. Utterance ``Ses04M\_impro02\_F024'' from the IEMOCAP dataset has two ``angry'' labels and two ``frustrated'' labels as shown in Figure~\ref{fig: iemo-22}. The EDL system predicts this utterance as ``frustrated'' with a belief mass of 0.567 and an overall uncertainty score of 0.433, which reveals that the system fails to detect the utterance as NMA. 
A possible cause of this failure is that the model gets confused by MA utterances seen in the training that convey similar emotional content, such as  ``Ses05M\_impro01\_M014'' whose annotations are shown in Figure~\ref{fig: iemo-12} with an MA emotion class ``frustrated''. Although one annotator considered it as ``angry'', the MA ground-truth target was ``frustrated'' in a classification-based system.
Both utterances occur within a dyadic situation where two people disagree, with the speaker being the one who compromises, feeling unhappy and frustrated. Such similar emotional content may confuse a classification-based system to also predict the NMA utterance as frustrated. 
It is worth noting that data with the same distribution as ``Ses04M\_impro02\_F024'', which has tied votes, is not included during the training of a classification-based model because there is no majority vote available to serve as ground truth.

\begin{figure}[H]
    \centering
    \vspace{-1ex}
    \includegraphics[width=0.95\linewidth]{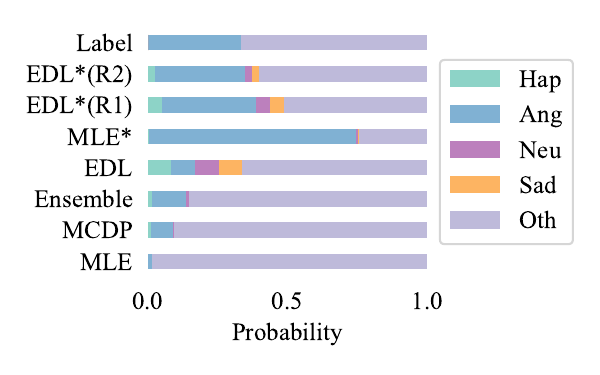}
    \vspace{-2ex}
    \caption{Predicted emotion distribution of MA utterance ``Ses05M\_impro01\_M014''.}
    \label{fig: iemo-12-distr}
\end{figure}

This complex emotional expression can be better described by the distribution-based EDL* systems. The predicted distribution of the MA utterance is shown in Figure~\ref{fig: iemo-12-distr} and the predicted distribution of the NMA utterance can be found in Figure~\ref{fig: case study}(a). It can be seen that the classification-based methods produce a similar distribution for the two utterances, with ``frustrated'' being dominant. However, the proposed EDL* methods can better match the label distribution and distinguish between these two cases. Although not been trained on NMA data, the EDL* methods are still capable of providing accurate predictions of its emotional content. This is a key benefit of the distribution-based approaches.

\begin{figure}[H]
    \centering
    \vspace{-1ex}
    \includegraphics[width=0.8\linewidth]{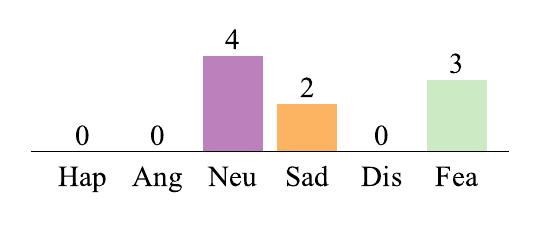}
    \vspace{-3ex}
    \caption{Human annotations for MA utterance ``1087\_IEO\_FEA\_LO''.}
    \vspace{-1ex}
    \label{fig: cremad-423}
\end{figure}

\begin{figure}[H]
    \centering
    \vspace{-3ex}
    \includegraphics[width=0.8\linewidth]{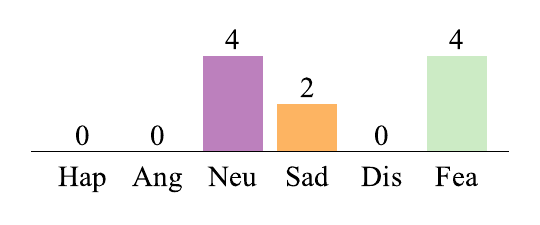}
    \vspace{-3ex}
    \caption{Human annotations for NMA utterance ``1052\_ITH\_FEA\_XX''.}
    \vspace{-1ex}
    \label{fig: cremad-424}
\end{figure}

\begin{figure}[H]
    \centering
    \includegraphics[width=0.95\linewidth]{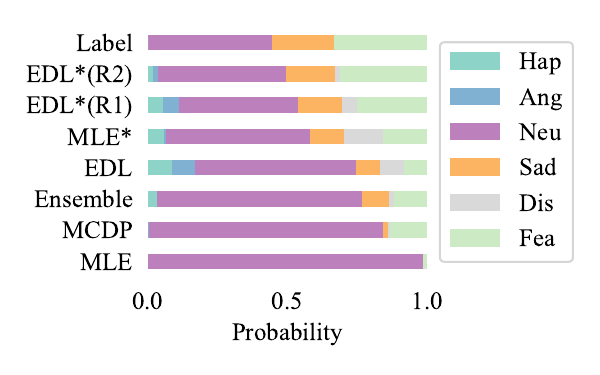}
    \vspace{-2ex}
    \caption{Predicted emotion distribution of MA utterance ``1087\_IEO\_FEA\_LO''.}
    \label{fig: cremad-43-distr}
\end{figure}

Next, we provide a typical false positive instance where an MA utterance is mis-detected as OOD. The MA utterance ``1087\_IEO\_FEA\_LO'' from the CREMA-D dataset has four ``neutral'', two ``sad'', and three ``fear'' human labels as in Figure~\ref{fig: cremad-423}. The NMA utterance ``1052\_ITH\_FEA\_XX'' has four ``neutral'', two ``sad'', and four ``fear'' human labels as in Figure~\ref{fig: cremad-424}. The OOD system successfully predicts the NMA utterance as OOD with an overall uncertainty of 0.691 while also predicting the MA utterance as an OOD sample with an overall uncertainty of 0.623.\footnote{Assume the OOD detection threshold is taken as 0.5.} This failure is possible because the MA utterance ``1087\_IEO\_FEA\_LO'' contains a complex mixture of emotions shown by the rather flat label distribution similar to ``1052\_ITH\_FEA\_XX'', which confuses the OOD detection system. Note that the MA class ``neutral'' in Figure~\ref{fig: cremad-423} comprises only $\frac{4}{4+2+3}\times100\%=44.4\%$ of the annotations and hence is not an absolute majority, which reduces the severity of this detection error.
Again, the distribution-based EDL* methods show superior capability in handling such complex cases. The predicted distribution of the MA utterance is shown in Figure~\ref{fig: cremad-43-distr} and the predicted distribution of the NMA utterance can be found in Figure~\ref{fig: cremad-case}(b). For the MA utterance, although human opinions diverge, the classification-based methods only capture the majority prediction, with the predicted distribution being dominated by ``neutral''. However, the emotion distribution predicted by the proposed EDL* methods retains the probability for ``sad'' and ``fear'' which accounts for the minority human opinions. Therefore, we show that the proposed EDL* method improves over the OOD system by providing a more comprehensive representation of emotional content as well as a more inclusive representation of human opinions.

\end{document}